\documentclass{Interspeech}

\usepackage{algorithmic}
\usepackage{algorithm}
\usepackage{xcolor}
\usepackage{cite}
\usepackage{subcaption}

\interspeechcameraready

\title{Scheduled Interleaved Speech-Text Training for Speech-to-Speech Translation with LLMs}

\author[affiliation={1}]{Hayato}{Futami}
\author[affiliation={1}]{Emiru}{Tsunoo}
\author[affiliation={1}]{Yosuke}{Kashiwagi}
\author[affiliation={1}]{Yuki}{Ito}
\author[affiliation={2}]{Hassan}{Shahmohammadi}
\author[affiliation={3}]{Siddhant}{Arora}
\author[affiliation={3}]{Shinji}{Watanabe}
\affiliation{}{Sony Group Corporation}{Japan}
\affiliation{}{Sony Europe}{Germany}
\affiliation{}{Carnegie Mellon University}{USA}

\email{Hayato.Futami@sony.com}

\usepackage{comment}

\begin{document}

\maketitle

\begin{abstract}
Speech-to-speech translation (S2ST) has been advanced with large language models (LLMs), which are fine-tuned on discrete speech units.
In such approaches, modality adaptation from text to speech has been an issue.
LLMs are trained on text-only data, which presents challenges to adapt them to speech modality with limited speech-to-speech data.
To address the training difficulty, we propose scheduled interleaved speech--text training in this study.
We use interleaved speech--text units instead of speech units during training, where aligned text tokens are interleaved at the word level.
We gradually decrease the ratio of text as training progresses, to facilitate progressive modality adaptation from text to speech.
We conduct experimental evaluations by fine-tuning LLaMA3.2-1B for S2ST on the CVSS dataset.
We show that the proposed method consistently improves the translation performances, especially for languages with limited training data.
\end{abstract}

\section{Introduction}
Speech-to-speech translation (S2ST) is a promising technology converts speech from one language to another, which can be useful for communicating with people who do not share the same language.
Traditionally, S2ST has been solved by cascaded approaches, which consist of automatic speech recognition (ASR), machine translation (MT), and text-to-speech (TTS) synthesis \cite{Nakamura06-ATR}.
In contrast, end-to-end S2ST systems have emerged \cite{Jia19-DSST, Jia21-Translatotron2, Lee22-DSST, Inaguma23-UnitY, Seamless23-SeamlessM4T, Rubenstein23-AP, Dong24-Polyvoice, Peng24-MSLMS2ST}, which are jointly optimized to translate speech in a source language into speech in a target language.
They have the advantage of simplified architecture and avoiding error propagation through joint optimization.

Recently, the end-to-end systems have adopted discrete speech units, instead of continuous mel-spectrogram features \cite{Lee22-DSST, Inaguma23-UnitY, Seamless23-SeamlessM4T, Rubenstein23-AP, Dong24-Polyvoice, Peng24-MSLMS2ST}.
There are mainly two types of discrete units: semantic and acoustic units.
Semantic units capture semantic information in speech and are obtained from self-supervised (SSL) speech encoders \cite{Hsu21-HuBERT, Chung21-W2vb}.
Acoustic units capture acoustic details and are obtained from neural codecs \cite{Zeghidour21-SS, Defossez22-EC}.
For S2ST, semantic speech LMs are used for semantic translation, and acoustic LMs are connected for high-quality speech generation, often with preserving speaker identities \cite{Rubenstein23-AP, Dong24-Polyvoice, Peng24-MSLMS2ST}.

After the success of LLMs in the natural language processing field \cite{Brown20-GPT3}, various efforts have been made to fine-tune pre-trained LLMs for speech processing \cite{Fathullah23-PLLM, Wang23-SLM, Wu23-ODA, Maiti23-VoxtLM, Chu23-QwenAudio, Zhang23-SGPT, Tang24-SALMONN}.
They have been mainly applied to ASR and speech-to-text translation tasks.
The use of discrete speech units paves the way for S2ST based on pre-trained LLMs.
AudioPaLM \cite{Rubenstein23-AP}, a representative study on this, fine-tunes a text LLM (PaLM2 \cite{Anil23-PaLM2}) on semantic units.
Due to strong language understanding capabilities of the LLM, AudioPaLM outperforms existing speech-to-text and speech-to-speech translation systems in a wide range of languages.
In this study, we build an S2ST system following AudioPaLM.

In fine-tuning text LLMs for the speech tasks, modality adaptation has been an issue.
As LLMs are trained on large-scale text-only data, it would be challenging to adapt LLMs to the speech domain with limited supervised data.
There are length and representation gaps between speech and text modalities \cite{Liu20-BMG, Tsiamas24-PLZ}, which have been addressed by length compression \cite{Wu23-ODA, Tsiamas24-PLZ}, modality adapter modules \cite{Tang24-SALMONN}, and training objectives \cite{Tsiamas24-PLZ}.
In AudioPaLM \cite{Rubenstein23-AP}, an SSL speech encoder is fine-tuned with an ASR task before LLM fine-tuning.
This helps mitigate the modality gap by obtaining discrete speech units that are similar to the text representations, leading to substantial performance improvement.

More recently, SpiRitLM \cite{Nguyen24-SLM} has been proposed as a speech--text multimodal LM, which is pre-trained on large-scale speech-only, text-only, and aligned speech--text data.
SpiRitLM employs interleaved speech--text representations for training on aligned data, where speech and aligned text units are randomly mixed at the word level.
The training is reported to achieve better modality transfer, which leads to performance improvement on speech and text comprehension tasks.

\begin{figure}[t]
\centering
\includegraphics[width=0.99\columnwidth]{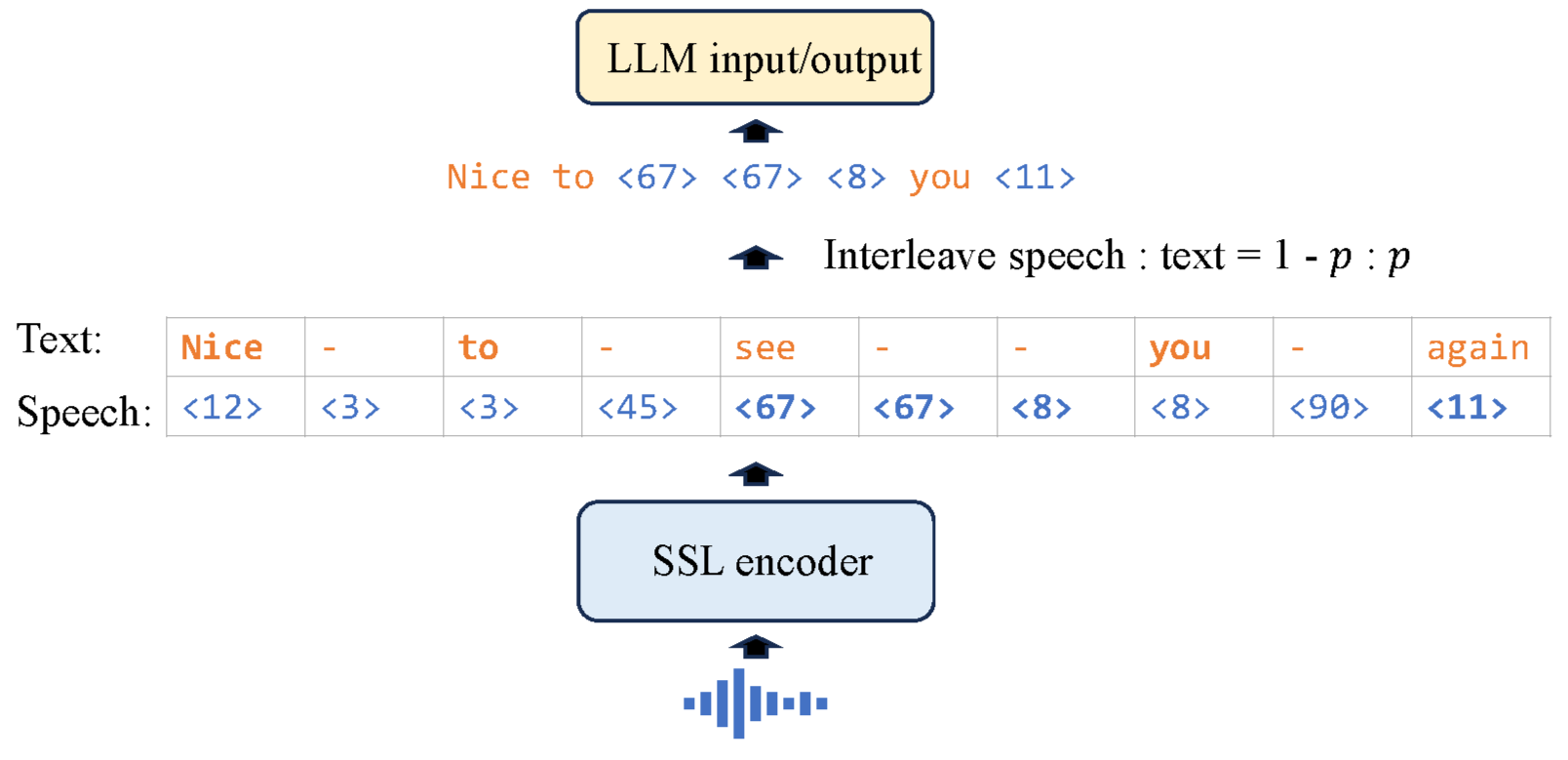}
\caption{Interleaved speech--text units used as input and output of LLM. Text ratio $p$ is gradually decreased as training progresses, which we call scheduled interleaved training.}
\label{fig:proposed}
\vspace{-10pt}
\end{figure}

In this study, we propose scheduled interleaved speech--text training, shown in Figure \ref{fig:proposed}.
We focus on {\it fine-tuning} pre-trained LLMs with limited supervised data for S2ST, instead of {\it pre-training foundational} speech--text LMs as in \cite{Nguyen24-SLM}.
Since pre-trained LLMs do not have the capabilities to interpret and generate speech units, we propose progressively adapting LLMs to speech units.
We use word-level interleaved speech and text units for both the input and output ends in S2ST.
For progressive adaptation, we propose gradually decaying the ratio of text units to zero, which finally results in all being speech units as expected during inference.

\begin{figure}[t]
\centering
\includegraphics[width=0.8\columnwidth]{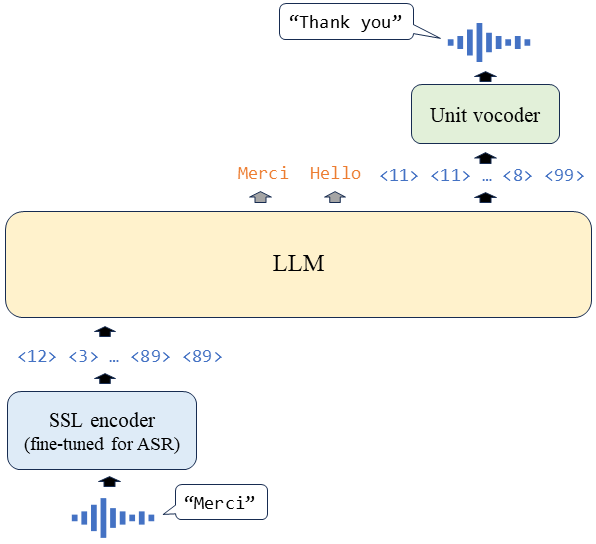}
\caption{Our LLM-based speech-to-speech translation system.}
\label{fig:system}
\vspace{-10pt}
\end{figure}

In experimental evaluations, we built an S2ST system by fine-tuning LLaMA3.2-1B LLM \cite{Dubey24-Llama3} on the CVSS corpus \cite{Jia22-CV}.
We used discrete speech units from a w2v-BERT encoder, which was fine-tuned for ASR with the CTC objective.
For waveform generation, we used a unit-based HiFi-GAN vocoder \cite{Polyak21-SR, Kong20-HFG}.
The overall system is shown in Figure \ref{fig:system}.
We show that the proposed scheduled interleaved speech--text training improved the S2ST performances, which can be attributed to better modality adaptation.
The proposed method was consistently effective across languages, and was especially effective for languages with limited training data.

\section{Related work}

\subsection{Speech-to-speech translation (S2ST)}
End-to-end speech-to-speech translation (S2ST) systems have been actively studied, which is jointly optimized as a speech-to-speech task \cite{Jia19-DSST, Jia21-Translatotron2, Lee22-DSST, Inaguma23-UnitY, Seamless23-SeamlessM4T, Rubenstein23-AP, Dong24-Polyvoice, Peng24-MSLMS2ST}.
Early studies have solved the task by directly predicting spectrograms in a target language with sequence-to-sequence models \cite{Jia19-DSST, Jia21-Translatotron2}.
Recently, speech synthesis has emerged to be conducted with autoregressive LMs on discrete speech units \cite{Polyak21-SR, Lakhotia21-GSLM, Borsos23-ALM, Wang23-VALLE, Zhang23-VALLEX}.
Two types of discrete speech units have been considered.
The first type is semantic units, which represent semantic information obtained from SSL speech encoders.
The second type is acoustic units, which represent acoustic details from neural codecs.
Generative spoken language modeling (GSLM) is trained on semantic units for speech continuation \cite{Lakhotia21-GSLM}.
AudioLM is trained on both semantic and acoustic units \cite{Borsos23-ALM}.
VALL-E extends LMs trained on acoustic units for zero-shot TTS \cite{Wang23-VALLE, Zhang23-VALLEX}.
% , and is further extended for cross-lingual voice transfer (VALL-E X) \cite{Zhang23-VALLEX}.
Following this trend, S2ST has also been solved with discrete speech units.
% VALL-E X realizes cross-lingual zero-shot TTS
UnitY \cite{Inaguma23-UnitY} adopts semantic units with two-pass architecture of speech-to-text and text-to-unit conversion, which are also trained on massively multilingual data, known as SeamlessM4T \cite{Seamless23-SeamlessM4T}.
AudioPaLM fine-tunes a pre-trained LLM on semantic units, and generate speech via acoustic units by AudioLM \cite{Borsos23-ALM}.
PolyVoice \cite{Dong24-Polyvoice} and MSLM-S2ST \cite{Peng24-MSLMS2ST} adopt the similar architecture that consists of semantic and acoustic LMs.
In this study, we focus on how to fine-tune a pre-trained LLM and semantic accuracy in translation.
As in AudioPaLM, we focus on fine-tuning a pre-trained LLM on semantic units, and generating speech by unit HiFi-GAN vocoder, as in \cite{Seamless23-SeamlessM4T, Polyak21-SR} \footnote{Instead of AudioLM, a unit vocoder is used due to its simplicity and training stability}.

\subsection{LLM for speech processing}
The NLP field has witnessed significant progress with LLMs \cite{Brown20-GPT3, Anil23-PaLM2, Dubey24-Llama3}.
Various efforts have been conducted to extend pre-trained LLMs to the speech modality, in the context of multi-modal LLMs \cite{Fathullah23-PLLM, Wang23-SLM, Wu23-ODA, Maiti23-VoxtLM, Chu23-QwenAudio, Zhang23-SGPT, Tang24-SALMONN}.
% For this purpose, we use speech encoder as a frontend of LLM.
They add a speech encoder as a frontend to the LLM to enable speech input, where foundational SSL or speech recognition models are adopted.
In addition, length and modality adapters are often introduced to fill the modality gap between the speech encoder and the LLM \cite{Wu23-ODA, Tang24-SALMONN}.
These models use continuous features as input to the LLM and focus mainly on speech-to-text tasks.
Instead of continuous representations, several approaches leverage discrete speech units and integrate speech generation tasks \cite{Zhang23-SGPT, Maiti23-VoxtLM}, including S2ST \cite{Rubenstein23-AP}.
Recently, SpiRitLM has been proposed, which is pre-trained on interleaved speech--text, and shows strong speech--text comprehension and in-context learning capabilities \cite{Nguyen24-SLM}. 
This is also extended with synthetic interleaved data \cite{Zeng24-SSTP}.
Instead of pre-training foundational speech--text LMs that can freely mix speech and text, we focus on fine-tuning a pre-trained LLMs with limited supervised data and propose a scheduling that gradually decreases the text ratio.

\section{Method}
\label{sec:method}

\subsection{Speech-to-speech translation system}

As shown in Figure \ref{fig:system}, we adopt a speech-to-speech translation (S2ST) system fine-tuned from an LLM, as in AudioPaLM \cite{Rubenstein23-AP}.
Let $S^{\rm src}$ denote semantic units in a source language, and $S^{\rm tgt}$ denote those in a target language.
End-to-end S2ST is trained to predict $S^{\rm tgt}$ from $S^{\rm src}$, by minimizing the following objective function:
\begin{align}
\mathcal{L} = - \log p(S^{\rm tgt} | S^{\rm src} ; \theta),
\end{align}
where $\theta$ denotes the model parameter.
As done in \cite{Rubenstein23-AP}, we consider predicting transcribed text $T^{\rm src}$ and translated text $T^{\rm tgt}$ as intermediate steps, which is jointly modeled as:
\begin{multline}
\label{eq:baseline}
\mathcal{L} = - \log p(S^{\rm tgt} | T^{\rm tgt}, T^{\rm src}, S^{\rm src} ; \theta) \\
p(T^{\rm tgt} | T^{\rm src}, S^{\rm src} ; \theta) p(T^{\rm src} | S^{\rm src} ; \theta),
\end{multline}
which we call chain-of-thought (CoT) prompting.
Note that the model has access to $S^{\rm src}$ when predicting $S^{\rm tgt}$, enabling end-to-end optimization \cite{Rubenstein23-AP}.
The model parameter is initialized with that of text LLM.
Its vocabulary $\mathcal{V}$ is extended to a combination of the vocabulary of speech units $\mathcal{V}_s$ and that of text BPEs $\mathcal{V}_t$ from the LLM, i.e. $\mathcal{V} = \mathcal{V}_s \cup \mathcal{V}_t$.

LLMs have been pre-trained solely on text tokens, without seeing speech data.
Therefore, it would be difficult for LLMs to adapt to speech units in the fine-tuning stage, as supervised training data is often limited.
There are modality gaps between speech and text in terms of their lengths and representations \cite{Liu20-BMG}, which makes the cross-modal fine-tuning difficult.
It has been demonstrated that fine-tuning an SSL encoder for ASR is effective \cite{Rubenstein23-AP}, which would reduce the modality gap and alleviate the training difficulty.

\subsection{Scheduled interleaved speech--text training}

In addition to ASR fine-tuning of SSL, we propose scheduled interleaved speech--text training inspired by SpiRitLM \cite{Nguyen24-SLM}, for better modality adaptation.
First, we calculate the word-level alignment of speech.
As we adopt a speech encoder fine-tuned for CTC-based ASR, we can exploit the alignment given by the CTC \cite{Kurzinger20-CTC}, without relying on any external aligners.
Then, we randomly mix speech and aligned text units at word level, to construct the interleaved speech--text units, as shown in Figure \ref{fig:proposed}.
We use the interleaved speech--text units $I^{\rm src}_p$ and $I^{\rm tgt}_p$ instead of speech semantic units during LLM fine-tuning, where $p$ denotes the text ratio.
The Eq.(\ref{eq:baseline}) is rewritten with $I^{\rm src}_p$ and $I^{\rm tgt}_p$ as:
\begin{multline}
\mathcal{L} = - \log p(I^{\rm tgt}_p | T^{\rm tgt}, T^{\rm src}, I^{\rm src}_p ; \theta) \\
p(T^{\rm tgt} | T^{\rm src}, I^{\rm src}_p ; \theta) p(T^{\rm src} | I^{\rm src}_p ; \theta).
\label{eq:chain-of-thought}
\end{multline}

Algorithm \ref{algo:interleaved} explains the detail of the interleaving algorithm.
The inputs are speech semantic units $S$ ($S^{\rm src}$ or $S^{\rm tgt}$) and corresponding word alignments $A$, represented as start and end frame indices $a_i, b_i$ in $S$, and word representations $w_i$.
The output is interleaved speech--text units $I_p$ ($I^{\rm src}_p$ or $I^{\rm tgt}_p$).
We randomly select word spans, by selecting start index $j$ from remaining indices $J$ (Line 3) and span length $l$ from a Poisson distribution (Line 4).
Speech units in the selected span $s_{a_j}, ..., s_{b_{j+l}}$ are replaced with corresponding Byte Pair Encoding (BPE) tokens BPE($w_j, ..., w_{j+l}$) (Line 5).
The text replacement is applied while the percentage of words selected does not exceed the text ratio $p$ (Line 2).

\begin{figure}[t]
\vspace{-10pt}
\begin{algorithm}[H]
\caption{Interleaved speech--text units}
\hspace*{\algorithmicindent} \textbf{Input:} speech semantic units $S = \{s_1, ..., s_M\}$, word alignments $A = \{(a_1,b_1,w_1),...,(a_j,b_j,w_j),...,(a_N,b_N,w_N)\}$ \\
\hspace*{\algorithmicindent} \textbf{Output:} interleaved speech--text units $I_p$ \\
\label{algo:interleaved}
\begin{algorithmic}[1]
\STATE Initialize $I_p \leftarrow S, J \leftarrow \{0, ..., N\}$.
\WHILE{$N - |J| \leq p \cdot N$}
\STATE Randomly select $j \in J$.
\STATE Sample length $l \sim {\rm Poission}(\lambda)$.
\STATE Replace $s_{a_j}, ..., s_{b_{j+l}}$ in $I_p$ with BPE($w_j, ..., w_{j+l}$).
\STATE $J \leftarrow J - \{ j, ..., j+l \}$.
\ENDWHILE
%\UNTIL{$N - |J| \gt p \cdot N$}
\end{algorithmic}
\end{algorithm}
\vspace{-20pt}
\end{figure}

In this study, we consider scheduling of the text interleaving ratio $p$, according to training steps.
We propose decaying the text ratio $p$ as training progresses (e.g. every $300$ steps), to achieve seamless modality shift from a text LLM to a speech LLM.
This means that the ratio of speech units, longer than text units, are gradually increased.
Its length and learned representations are gradually adapted, as shown later in Figure \ref{fig:analysis}, which would mitigate the difficulty of modality adaptation.
The scheduling of $p$ is the main difference from SpiRitLM \cite{Nguyen24-SLM}.
In addition, we focus on fine-tuning for the S2ST task with limited supervised data, rather than large-scale pre-training as in SpiRitLM.
Different from SpiRitLM, we also investigate multilingual settings, semantic units with ASR fine-tuning, and a combination with chain-of-thought prompting.

\begingroup
\renewcommand{\arraystretch}{1.2}
\begin{table}[t]
\caption{The effect of ASR fine-tuning (FT) of w2v-BERT (W2VB) and Chain-of-Thought (CoT) prompting.}
\label{tab:baseline}
\centering
\begin{tabular}{lc} \hline
 & Fr-en \\
 & ASR-BLEU($\uparrow$) / UTMOS($\uparrow$) \\ \hline
W2VB & $9.6/4.35$ \\
ASR FT W2VB (baseline) & $\bm{28.8}/4.21$ \\
w/o CoT & $13.7/4.21$ \\ \hline
\end{tabular}
\vspace{-10pt}
\end{table}
\endgroup

\begingroup
\renewcommand{\arraystretch}{1.2}
\begin{table*}[t]
\caption{Speech-to-speech translation (S2ST) performances on CVSS, reported with ASR-BLEU and UTMOS scores. Proposed scheduled interleaved speech-text training (ILT) outperformed baseline without it in ASR-BLEU.}
\label{tab:speech-to-speech-translation}
\centering
\begin{tabular}{lccccccc} \hline
 & Fr-en & De-en & Es-en & Ca-en & It-en & Ru-en & Pt-en \\
 & ($180$h) & ($119$h) & ($97$h) & ($81$h) & ($28$h) & ($16$h) & ($7$h) \\ \hline
Baseline & $28.8/4.21$ & $27.3/4.22$ & $33.5/4.28$ & $23.8/4.17$ & $12.8/4.23$ & $6.0/4.25$ & $10.3/4.17$ \\
+Scheduled ILT & $\bm{29.5}/4.23$ & $\bm{29.2}/4.21$ & $\bm{33.9}/4.28$ & $\bm{25.9}/4.21$ & $\bm{19.5}/4.23$ & $\bm{14.1}/4.23$ & $\bm{19.5}/4.17$ \\
+ILT w/o scheduling & $25.8/4.11$ & $22.1/4.11$ & $21.9/4.29$ & $17.8/4.10$ & $14.4/4.22$ & $6.0/4.12$ & $12.0/4.11$ \\ \hline
\end{tabular}
\vspace{-10pt}
\end{table*}
\endgroup

\begingroup
\renewcommand{\arraystretch}{1.2}
\begin{table}[t]
\caption{Ablation studies of S2ST on CVSS Pt-en, reported with ASR-BLEU and UTMOS scores.}
\label{tab:ablation}
\centering
\begin{tabular}{lcccc} \hline
 & Pt-en \\ \hline
Scheduled ILT & $\bm{19.5}/4.17$ \\
- Interleave input only & $11.1/4.18$ \\
- Interleave output only & $12.6/4.18$ \\
- Mask instead of interleave & $6.2/4.17$ \\
- Interleave without alignment & $11.5/4.16$ \\ \hline
\end{tabular}
\vspace{-10pt}
\end{table}
\endgroup

\section{Experimental evaluations}
We conducted speech-to-speech translation (S2ST) experiments based on the CVSS \cite{Jia22-CV} corpus.
The CVSS corpus is a widely used corpus for multilingual S2ST, built by speech synthesis from the CoVoST2 \cite{Wang21-CV} speech-to-text translation corpus.
CVSS consists of two versions: CVSS-C with a single high-quality canonical voice and CVSS-T with voice transfer from source speakers.
We used the CVSS-C in the experiments.

As described in Section \ref{sec:method}, we used semantic units from the fine-tuned SSL encoder.
We fine-tuned the w2v-BERT encoder for ASR on all the CoVoST2 and CVSS training data ($21$ languages plus English).
The w2v-BERT fine-tuning was done with the Adam optimizer with a learning rate of $2$e$-5$.
We added a linear layer for BPE of size $5$K, and trained the entire model using the CTC objective.
Then, we trained $k$-means of cluster size $2048$ on features from the $20$th w2v-BERT layer, using the CoVoST+CVSS data.
We used cluster indices from the $k$-means as semantic units, without applying deduplication.
We fine-tuned LLaMA3.2-1B\footnote{https://huggingface.co/meta-llama/Llama-3.2-1B} LM on each language subset of CVSS.
The vocabulary of the LM jointly consists of around $12$K text BPEs (identical to the original LLaMA) and $2048$ speech units.
The LM fine-tuning was done with Adam optimizer of learning rate of $5$e$-5$ and dropout rate of $0.2$.
During training and inference, we applied CoT prompting in Eq. (\ref{eq:chain-of-thought}), where speech recognition and text translation are conducted as intermediate steps.
We implemented the w2v-BERT and LM fine-tuning with the ESPnet toolkit \cite{Watanabe18-ESP}.
For waveform generation, we trained a unit HiFi-GAN vocoder on the CVSS data, which consists of synthesized single-speaker speech in English.
In the unit HiFi-GAN, we used semantic units from w2v-BERT as input, instead of mel-spectrograms.
We implemented it with the ParallelWaveGAN toolkit following its LJSpeech recipe \footnote{https://github.com/kan-bayashi/ParallelWaveGAN}.

First, we tested the effectiveness of ASR fine-tuning of w2v-BERT and CoT prompting in Table \ref{tab:baseline}.
We used the CVSS French-to-English (Fr-en).
We fine-tuned the LM on the train set of Fr-en, and evaluated it on the test set \footnote{We randomly sampled $1$K samples from the test set. The number of samples is too large (more than $10$K) in some languages.}.
The translation accuracy was evaluated with ASR-BLEU scores, where BLEU scores were calculated on ASR transcripts of generated speech, by Whisper \cite{Radford22-Whisper} large-v3 model.
The audio quality was evaluated with UTMOS \cite{Saeki22-UTMOS}.
We confirmed that ASR fine-tuning (ASR FT) was crucial for the accuracy, as observed in \cite{Rubenstein23-AP, Huang24-IDLLM}.
We also confirmed that CoT prompting was effective, as in \cite{Rubenstein23-AP}.
Hereafter, we regard the LM applied both of them as a baseline.

Table \ref{tab:speech-to-speech-translation} shows the main experiments for proposed scheduled interleaved training (ILT), on subsets of seven languages in CVSS.
We compared three training methods: the baseline without any interleaving (Baseline), proposed scheduled interleaved speech--text training (Scheduled ILT), and ILT without scheduling (ILT w/o scheduling).
For ILT, we obtained the word alignments via CTC segmentation \cite{Kurzinger20-CTC}, by using the fine-tuned w2v-BERT encoder.
In scheduled ILT, the text ratio $p$ in Algorithm \ref{algo:interleaved} was initialized with $0.9$.
The text ratio $p$ was gradually decreased by $0.1$, every $300$ training steps.
In ILT without scheduling, we used a constant $p$ of $0.3$.
Table \ref{tab:speech-to-speech-translation} shows that scheduled ILT outperformed the baseline in terms of translation accuracy (ASR-BLEU) in all the seven languages, demonstrating the effectiveness of our proposed method.
We also found that the improvements were particularly large in the languages with less training data (It-en, Ru-en, and Pt-en).
It would be difficult to adapt to the speech modality with limited speech-to-speech data, but the proposed training alleviated the difficulty.
This is favorable for real-world applications, where supervised training data is often limited.
Also, we confirmed that the qualities of generated speech (UTMOS) were in a high level, regardless of the training methods.
Finally, we compared scheduled ILT with one w/o scheduling, and found that the scheduling was effective for the translation performances.

Table \ref{tab:ablation} shows ablation studies on the components of scheduled ILT, on the CVSS Portuguese-to-English (Pt-en) subset.
As described in Section \ref{sec:method}, we used interleaved speech--text units at both the input (source) and output (target) sides $I^{\rm src}_p$ and $I^{\rm tgt}_p$.
To see whether interleaving in both sides is necessary, we compared this with interleaving in either input or output side (Row 2 or 3).
For example, interleaving in the input side only corresponds to using $S^{\rm tgt}$ instead of $I^{\rm tgt}_p$ in Eq. (\ref{eq:chain-of-thought}).
We show that interleaving in both sides was necessary for translation performances, comparing it with interleaving in the input or output side only.
Also, we see whether our method is not just a data augmentation method with noisy input and output, comparing it with masking instead of interleaving the text tokens (Row 4).
Specifically, we replaced selected speech units $s_{a_j}, ..., s_{b_{j+l}}$ with a special mask token (\url{[MASK]}) instead of BPE at Line 5 in Algorithm \ref{algo:interleaved}.
We observed that masking performed worse compared to speech--text interleaving.
Finally, to see whether word alignment is necessary, we apply interleaving without alignment (Row 5).
We assume that the words appear at equal intervals in speech.
This is represented as $a_i = i \cdot \lfloor M/N \rfloor $ and $b_i = (i+1) \cdot \lfloor M/N \rfloor - 1$ in Algorithm \ref{algo:interleaved}.
We observed that interleaving without alignment led to performance degradation, demonstrating the importance of alignment.

\begin{figure}[t]
\centering
\begin{minipage}[b]{0.49\columnwidth}
    \centering
    \includegraphics[width=0.99\columnwidth]{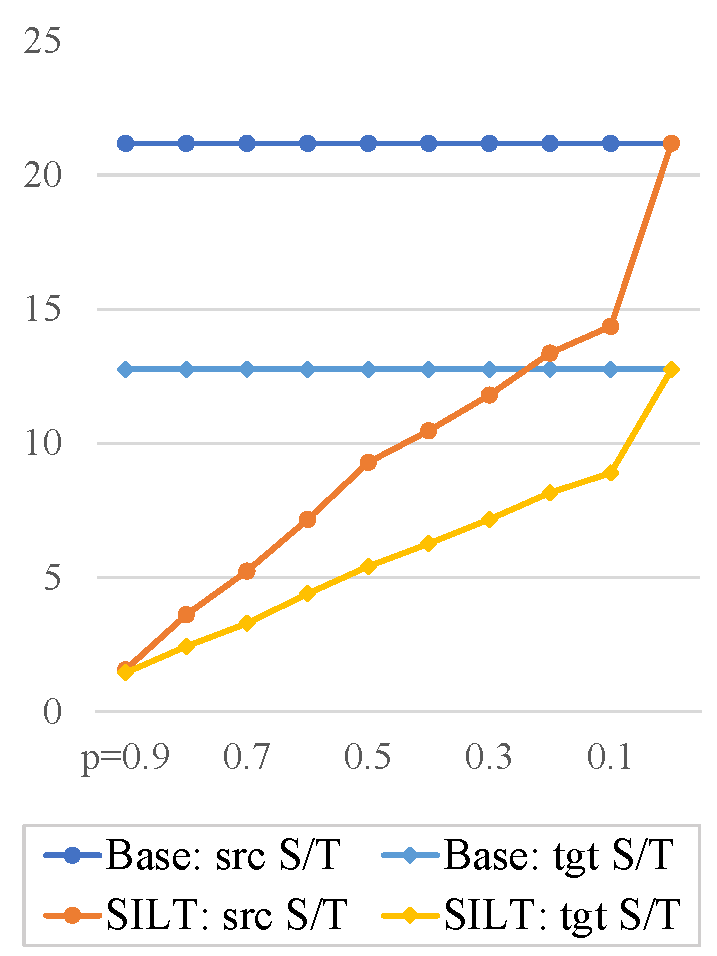}
    \subcaption{Length}
    \label{fig:a}
\end{minipage}
\begin{minipage}[b]{0.49\columnwidth}
    \centering
    \includegraphics[width=0.99\columnwidth]{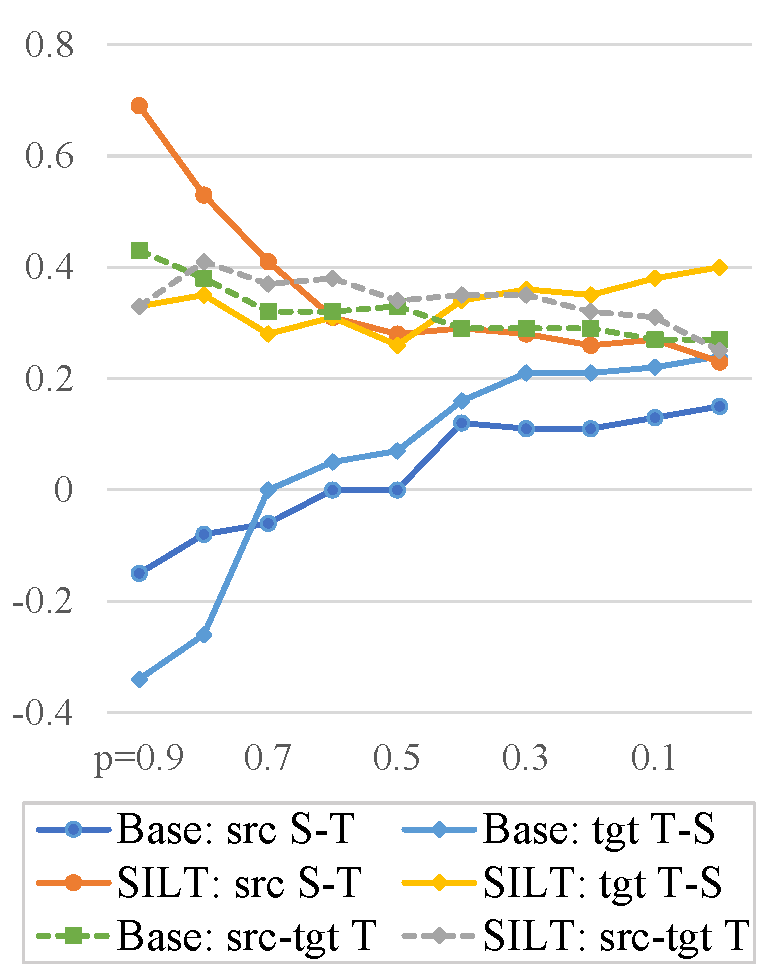}
    \subcaption{Representation similarity}
    \label{fig:b}
\end{minipage}
\caption{Length and learned representation gaps between speech (S) and text (T), seen during training of baseline (Base) and scheduled ILT (SILT). Text ratio $p$ is gradually decayed by $0.1$ every $300$ steps.}
\label{fig:analysis}
\vspace{-18pt}
\end{figure}

Figure \ref{fig:analysis} shows how proposed scheduled ILT affects the training in terms of length and learned representations between speech and text.
We used baseline and scheduled ILT models for Pt-En.
The left figure shows the ratio of the speech-unit length against the text-unit length i.e. $|S|/|T|$.
Instead of $S$, scheduled ILT adopts interleaved units $I_p$.
We measured it for both the source and target sides.
We found that the lengths significantly varied between speech and text, where the speech units are longer than text, by $21.7$ times in the source and $13.1$ times in the target sequence.
Scheduled ILT gradually decreased the text ratio $p$, so the lengths were gradually increased, as shown in Figure \ref{fig:analysis} \footnote{The jump on length was observed between $p=0.1$ and $0$. The actual text ratio is larger than $p$, because the loop in Algorithm \ref{algo:interleaved} was stopped when the text ratio exceeds $p$.}, which would reduce the training difficulty.
The right figure shows the cosine similarity of the hidden representations.
% Sum
We analyzed the hidden states of the last LLaMA layer, averaged over the sequence corresponding to each input type (e.g. $S^{\rm src}$).
We measured cosine similarity, between source speech and text (src S-T), between source and target text (src-tgt T), and between target text and speech (tgt T-S).
We found that the baseline model had difficulty in learning speech--text relationship in the early training stages.
On the other hand, in scheduled ILT, the speech--text similarities were much higher in the early stages, due to the high proportion of text in interleaving.
This would help learn the interaction between speech and text, and the learned representations at the final stage (without interleaving, $p=0$) were more similar than the baseline.
In contrast, scheduled ILT did not bring impact on text--text similarity, as our method improved speech--text adaptation.

\vspace{-5pt}
\section{Conclusions}
In this study, we have proposed a new training method named scheduled interleaved speech--text training for LLM-based speech-to-speech translation (S2ST).
We use interleaved speech--text units as the input and output of LLM, instead of the speech units, during fine-tuning LLM.
We gradually decrease the ratio of text in interleaving, for better modality adaptation from text to speech.
We have conducted S2ST experiments on the CVSS corpus, and have shown that our method consistently improved the translation performances across seven languages.
In future work, we will extend the proposed approach to other speech-to-speech systems, such as spoken dialog systems.

\bibliographystyle{IEEEtran}
\bibliography{mybib}

\end{document}